\documentclass[conference]{IEEEtran}
\usepackage{times}

\usepackage[numbers]{natbib}
\usepackage{multicol}
\usepackage[bookmarks=true]{hyperref}

\usepackage{amsmath,amssymb,bbm}
\usepackage{graphicx}
\usepackage{booktabs}
\usepackage{algorithm}
\usepackage{algpseudocode}

\usepackage{array,multirow}

\DeclareMathOperator*{\argmax}{arg\,max}

\usepackage{color}

\renewcommand{\baselinestretch}{.9636}
\makeatletter
\def\blfootnote{\xdef\@thefnmark{}\@footnotetext}
\makeatother

\begin{document}

\title{Shared Autonomy via Deep Reinforcement Learning}

\author{\authorblockN{Siddharth Reddy, Anca D. Dragan, Sergey Levine}
\authorblockA{Dept. of Electrical Engineering and Computer Science\\
University of California, Berkeley\\
\{sgr,anca,svlevine\}@berkeley.edu}}

\maketitle
\begin{abstract}
In shared autonomy, user input is combined with semi-autonomous control to achieve a common goal. The goal is often unknown ex-ante, so prior work enables agents to infer the goal from user input and assist with the task.
Such methods tend to assume some combination of knowledge of the dynamics of the environment, the user's policy given their goal, and the set of possible goals the user might target, which limits their application to real-world scenarios.
We propose a deep reinforcement learning framework for model-free shared autonomy that lifts these assumptions.
We use human-in-the-loop reinforcement learning with neural network function approximation to learn an end-to-end mapping from environmental observation and user input to agent action values, with task reward as the only form of supervision.
This approach poses the challenge of following user commands closely enough to provide the user with real-time action feedback and thereby ensure high-quality user input, but also deviating from the user's actions when they are suboptimal. We balance these two needs by discarding actions whose values fall below some threshold, then selecting the remaining action closest to the user's input.
Controlled studies with users ($n = 12$) and synthetic pilots playing a video game, and a pilot study with users ($n = 4$) flying a real quadrotor, demonstrate the ability of our algorithm to assist users with real-time control tasks in which the agent cannot directly access the user's private information through observations, but receives a reward signal and user input that both depend on the user's intent.
The agent learns to assist the user without access to this private information, implicitly inferring it from the user's input. This enables the assisted user to complete the task more effectively than the user or an autonomous agent could on their own.
This paper is a proof of concept that illustrates the potential for deep reinforcement learning to enable flexible and practical assistive systems.
\end{abstract}

\IEEEpeerreviewmaketitle

\blfootnote{See \url{https://sites.google.com/view/deep-assist} for supplementary materials, including videos and code.}

\section{Introduction}

Imagine the task of flying a quadrotor to a safe landing site. This problem is challenging for both humans and robots, but in different ways. For a human, controlling many degrees of freedom at once while dealing with unfamiliar quadrotor dynamics is hard. For a robot, understanding what makes a good landing location can be difficult, especially when the human has a future task in mind that might influence where they want the quadrotor to land now.

Shared autonomy~\cite{goertz1963manipulators,aigner1997human} aims to address this problem by combining user input with automated assistance. We focus on an area of shared autonomy in which information about the user's intent is hidden from the robot, in which prior work~\cite{muelling2017autonomy,javdani2015shared,perez2015fast,koppula2016anticipating,hauser2013recognition} has proposed approaches that infer the user's goal from their input and autonomously act to achieve it. These approaches tend to assume (1) a known dynamics model of the world, (2) a known goal representation (a set of possible goals), and (3) a known user policy given a goal.

\begin{figure}[t]
    \centering
    \includegraphics[width=\linewidth]{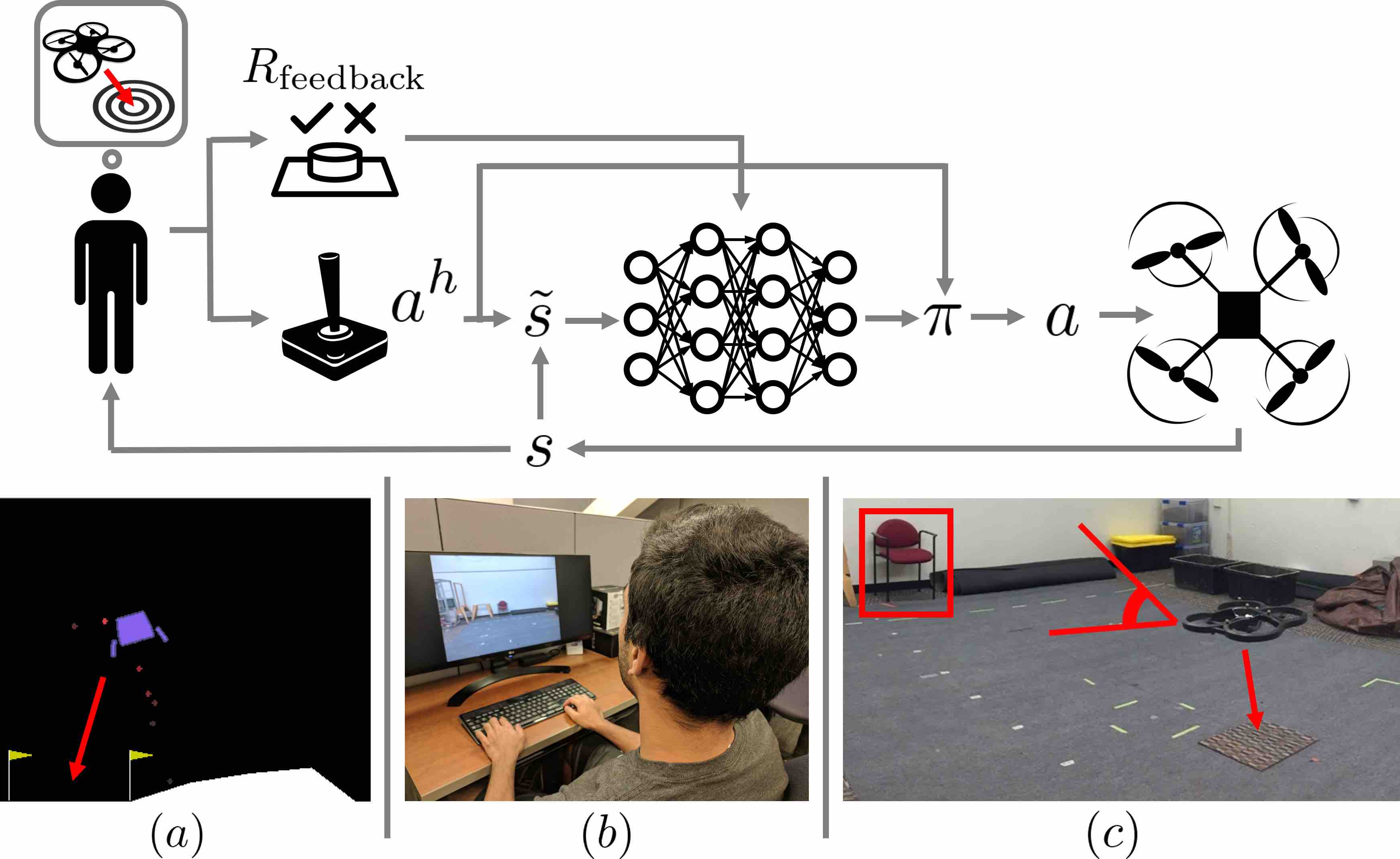}
    \caption{An overview of our method for assisting humans with real-time control tasks using model-free shared autonomy and deep reinforcement learning. We empirically evaluate our method on simulated pilots and real users playing the Lunar Lander game (a) and flying a quadrotor (b,c).}
    \label{fig:front-fig}
\end{figure}

For many real-world tasks, these assumptions constrain the adaptability and generality of the system. (1) Fitting an accurate global dynamics model can be more difficult than learning to perform the task. (2) Assuming a fixed representation of the user's goal (e.g., a discrete set of graspable objects) reduces the flexibility of the system to perform tasks in which the users' desires are difficult to specify but easy to evaluate (e.g., goal regions, or success defined directly on raw pixel input).
(3) User input can exhibit systematic suboptimality that prevents standard goal inference algorithms from recovering user intent by inverting a generative model of behavior.

Our goal is to devise a shared autonomy method that lifts these assumptions, and our primary contribution is a model-free deep reinforcement learning algorithm for shared autonomy that represents a step in this direction. The key idea is that training an end-to-end mapping from environmental observation and user input to agent action values, with task reward as the only form of supervision, removes the need for known dynamics, a particular goal representation, and even a user behavior model. From the agent's perspective, the user acts like a prior policy that can be fine-tuned, and an additional sensor generating observations from which the agent can implicitly decode the user's private information. From the user's perspective, the agent behaves like an adaptive interface that learns a personalized mapping from user commands to actions that maximizes task reward.

One of the core challenges in this work lies in adapting standard deep reinforcement learning techniques to leverage input from a human without significantly interfering in their real-time `feedback control loop' -- the user's ability to observe the consequences of their own actions, and adjust their inputs accordingly. Consistently ignoring the user's input can prevent them from using action feedback to improve the quality of their input.
To address this issue, we use human-in-the-loop deep Q-learning to learn an approximate state-action value function that computes the expected future return of an action given the current environmental observation and the user's control input. Rather than taking the highest-value action, our assistive agent executes the closest high-value action to the user's input, balancing the need to take optimal actions with the need to preserve the user's feedback control loop. This approach also enables the user to directly modulate the level of assistance through the parameter $\alpha \in [0, 1]$, which sets the threshold of the system's tolerance for suboptimal user actions.

Standard deep reinforcement learning algorithms pose another challenge for human-in-the-loop training: they typically require a large number of interactions with environment, which can be a burden on users. We approach this problem by decomposing the agent's reward function into two parts: known terms computed for every state, and a terminal reward provided by the user upon succeeding or failing at the task. This decomposition enables the system to learn efficiently from a dense reward signal that captures generally useful behaviors like not crashing, and also adapt to individual users through feedback. It also enables pretraining the agent in simulation without a user in the loop, then later fine-tuning -- instead of learning from scratch -- with user feedback. To further improve sample efficiency, our method is capable of incorporating inferred goals into the agent's observations when the goal space and user model are known.

We apply our method to two real-time assistive control problems: the Lunar Lander game and a quadrotor landing task
(see Figure \ref{fig:front-fig}).
Our studies with both human and simulated pilots suggest that our method can successfully improve pilot performance. We find that our method is capable of adapting to the unique types of suboptimality exhibited by different simulated pilots, and that by varying a hyperparameter that controls our agent's tolerance for suboptimal pilot controls, we are able to help simulated pilots who need different amounts of assistance. With human pilots, our method substantially improves task success and reduces catastrophic failure. Finally, we show that when the user policy or goal representation are known, our method can be combined with adaptations of existing techniques to exploit this knowledge.

\section{Related Work}

\noindent\textbf{Robotic teleoperation.} We build on shared autonomy work in which the system is initially unaware of the user's goal~\cite{green2008human,dragan2013policy,muelling2017autonomy,javdani2015shared,perez2015fast,koppula2016anticipating,hauser2013recognition} and explore problem statements with unknown dynamics, unknown user policy, and unknown goal representation.
The parallel autonomy~\cite{schwarting2017parallel} and outer-loop stabilization~\cite{broad2017learning} frameworks approach shared-control teleoperation from a different angle: instead of predicting user intent, they minimally adjust user input to achieve safe trajectories for tasks like semi-autonomous driving. Our agent's policy of executing a near-optimal action closest to the human's suggestion is inspired by this approach. Existing work in parallel autonomy requires
analytic descriptions of the environment, such as the explicit locations of road boundaries and a model of the behavior of other cars. Outer-loop stabilization requires knowledge of the user's goal. Our method is analogous, but for environments in which we do not have a dynamics model or a goal representation.

\noindent\textbf{Brain-computer interfaces.} A large body of work in brain-machine interfaces uses optimal control and reinforcement learning algorithms to implement closed-loop decoder adaptation~\cite{shanechi2016robust} for applications like prosthetic limb controllers that respond to neural signals from myoelectric sensors~\cite{pilarski2011online}. These algorithms typically track desired motion, whereas we focus on tasks with long-horizon goals.

\noindent\textbf{Reinforcement learning with human feedback.} Shared autonomy enables a semi-autonomous agent to interpret user input at test time. In contrast, human-in-the-loop reinforcement learning frameworks leverage human feedback to train autonomous agents that operate independently of the user at test time~\cite{warnell2017deep,knox2009interactively,knox2012reinforcement,lin2017explore}. These frameworks are applicable to settings where the agent has access to all task-relevant information (e.g., goals), but the reward function is initially unknown or training can be sped up by human guidance. We focus on the orthogonal setting where the agent does not have direct access to the information that is private to the user and relevant to the task, and will always need to leverage user input to accomplish the task; even after training. This is also the key difference between our method and inverse reinforcement learning~\cite{ng2000algorithms} and learning from demonstration~\cite{argall2009survey}, which generally require user interaction during training time but not at test time.

\noindent\textbf{Adaptive HCI.} While the bulk of the shared autonomy research discussed here exists in the context of the robotics literature, adaptive human-computer interfaces have been explored in computer graphics for animating virtual characters using motion capture data from humans~\cite{dontcheva2003layered}, in natural language processing for learning to act on natural language instructions from humans~\cite{Wang,branavan2009reinforcement}, and in formal methods for verification of semi-autonomous systems~\cite{seshia2015formal}. By not assuming a known user policy, our work also enables agents to adapt to a user's style of giving input.

\section{Background}

We first recap the reinforcement learning and shared autonomy problem statements on which we build in our method.

\subsection{Reinforcement Learning} \label{rl-background}

Consider a Markov decision process (MDP) with states $\mathcal{S}$, actions $\mathcal{A}$, transitions \mbox{$T : \mathcal{S} \times \mathcal{A} \times \mathcal{S} \to [0, 1]$}, reward function $R : \mathcal{S} \times \mathcal{A} \times \mathcal{S} \to \mathbb{R}$, and discount factor $\gamma \in [0, 1]$. In cases where the state is not fully observable, we can extend this definition to a partially-observable MDP (POMDP) in which there is an additional set of possible observations $\Omega$ and observation function $O : \mathcal{S} \times \Omega \to [0, 1]$. The expected future discounted return of taking action $a$ in state $s$ with policy $\pi : \mathcal{S} \times \mathcal{A} \to [0, 1]$ is expressed by the state-action value function $Q^{\pi}(s, a)$, and the goal in RL is to learn a policy $\pi^\ast$ that maximizes expected future discounted return.
One algorithm for solving this problem is Q-learning~\cite{watkins1992q}, which minimizes the Bellman error of the Q function,
\[ Q(s, a) - \gamma \mathbb{E}_{s' \sim T(\cdot \mid s, a)}\left[R(s, a, s') + \max_{a' \in \mathcal{A}} Q(s', a')\right], \]
as a proxy for maximizing return. We will build on this method to implement model-free shared autonomy.

\subsection{Shared Autonomy} \label{sa-background}

Prior work has formalized shared autonomy as a POMDP
\cite{javdani2015shared}.
The reward function, known to both the user and agent, depends on a goal $g \in \mathcal{G}$ known to the user but unknown to the agent. The set of candidate goals $\mathcal{G}$ is known to the agent.
The user follows a goal-conditioned policy $\pi_h : \mathcal{S} \times \mathcal{G} \times \mathcal{H} \to [0, 1]$ known to the agent, where $\mathcal{H}$ is the space of possible user inputs -- if the user suggests actions, then $\mathcal{H} = \mathcal{A}$.
The transition distribution $T$ is known to the agent. The agent's uncertainty in the goal can be formalized as partial observability, which leads to the following POMDP: the state space $\tilde{\mathcal{S}} = \mathcal{S} \times \mathcal{G}$ is augmented with the goal, the transition distribution $\tilde{T}((s_{t+1}, g) \mid s_t, g, a_t) = T(s_{t+1} \mid s_t, a_t)$ maintains a constant goal, and the observation distribution $O(s, a^h \mid s, g) = \pi_h(a^h \mid s, g)$ is given by the user policy where $a^h \in \mathcal{H}$ is the user input.
Prior work assumes the goal space $\mathcal{G}$, user policy $\pi_h$, and environment dynamics $T$ are known ex-ante to the agent, and solves the POMDP $(\tilde{\mathcal{S}}, \mathcal{A}, \tilde{T}, \tilde{R}, \mathcal{H}, O)$ using approximate methods like hindsight optimization~\cite{javdani2015shared}. In the following section, we introduce a different problem statement for shared autonomy which relaxes these assumptions.

\section{Model-Free Shared Autonomy}\label{mfsa}

We will relax the standard formulation in Section \ref{sa-background} to remove first the assumptions of known dynamics and the known observation model $\pi_h$ for the user's private information, and then the known set of candidate goals $\mathcal{G}$. We introduce a model-free deep reinforcement learning method, with variants that can also take advantage of a known observation model and goal space when they do exist, but still provide assistance even when they are not available.

\subsection{Problem Statement}
In our problem formulation, the transition $T$, the user's policy $\pi_h$, and the goal space $\mathcal{G}$ are no longer all necessarily known to the robot. The reward function, which still depends on the user's private information, is decomposed as:
\begin{equation} \label{rew-decomp-goal}
R(s, a, s') = \underbrace{R_{\text{general}}(s, a, s')}_\text{known} + \underbrace{R_{\text{feedback}}(s, a, s')}_\text{unknown, but observed}.
\end{equation}
This captures a structure typically present in shared autonomy: there are some terms in the reward that are known, such as the need to avoid collisions. We capture these in $R_{general}$. $R_{\text{feedback}}$ is a user-generated feedback that depends on their private information. We do not know this function. We merely assume the robot is informed when the user provides feedback (e.g., by pressing a button). In practice, the user might simply indicate once per trial whether the robot succeeded or not.

\noindent\textbf{Known-User-Policy: Unknown dynamics, known goal space and user policy.}
In this setting, the transition $T$ is unknown, but we have access to both $\mathcal{G}$ and the user's policy $\pi_h(a^h|s,g)$.
Having access to $\mathcal{G}$ structures $R_{\text{feedback}}$, which is now parameterized by the goal according to $R_{\text{feedback}}(s, a, s';g)$, and assigns high reward when $s'=g$, and 0 otherwise, without requiring manual indication from the user. We do not know $g$, but having access to $\pi_h$ enables us to infer $g$ via Bayesian inference.

\noindent\textbf{Known-Goal-Space: Unknown dynamics and user policy, known goal space.}
We also consider a version of the problem where we know $\mathcal{G}$, but do not make assumptions about the user's policy $\pi_h$. In this case, $R_{\text{feedback}}$ is still parameterized by the goal, but we must use a classification or regression model to predict the goal from the user's actions.

\noindent\textbf{Min-Assumptions: Unknown dynamics, user policy, and goal space.}
Most of our experiments will be concerned with this setting, where we no longer assume a goal representation. This provides us with a maximally general approach, where the user might imagine whichever goal they prefer, without the need to explicitly define the space of goals in advance. In this case, we do not know the functional form of $R_{\text{feedback}}$, nor do we assume any parameterization for it, we merely assume the robot can observe it (evaluate it) as it takes actions. This is typically a sparse terminal reward that signals whether the task was completed successfully, and comes from the user.

\subsection{Method Overview}

Our method takes observations of the environment and the user's controls or inferred goal (when available) as input, and produces a high value action or control output that is as close as possible to the user's control. We learn state-action values via Q-learning with neural network function approximation. In this section, we will describe how the agent combines user input with environmental observations, motivate and describe our choice of deep Q-learning for training the agent, and describe how the agent shares control with the user.

\subsection{Incorporating User Control}

Because we do not know dynamics in any of our problems of interest, we use a deep reinforcement learning agent which maps observations from its sensors to actions (or Q values for each action). We incorporate information from the user as useful observations for the agent.
Our method jointly embeds the agent's observation of the environment $s_t$ with the information from the user $u_t$ by simply concatenating them. The particular form of $u_t$ depends on the information that is available. Formally,
\begin{equation} \label{eq:general-embed}
\tilde{s}_t = \left[ \begin{array}{c} s_t \\ u_t \end{array} \right].
\end{equation}
When we do not know $\mathcal{G}$, we use the user's actions $a^h_t$ as $u_t$. When we know more about the possible user goals and policy, we set $u_t$ to the inferred goal $\hat{g}_t$.

\noindent\textbf{Known-User-Policy: Incorporating user control via Bayesian goal inference.} When the user's policy is available, it can be used to infer the maximum a posteriori estimate of the goal $\hat{g}_t$. We can instantiate Bayesian goal inference by using maximum entropy inverse reinforcement learning~\cite{ziebart2008maximum} with a goal-parameterized $Q$ function trained via Q-learning separately from our agent, analogously to prior work~\cite{javdani2015shared}. Each time step produces a better estimate of the goal $\hat{g}_t$, as additional actions reveal more about the user's intent.

\noindent\textbf{Known-Goal-Space: Incorporating user control via supervised goal prediction.} When we do not have a convenient model of the user's policy, we can use supervised prediction to compute the goal estimate $\hat{g}_t$. In this case, we use a separate recurrent LSTM network to predict the goal, conditioned on the sequence of states and user controls observed up to the current time $t$. Training data is collected from the user. As before, we concatenate $\hat{g}_t$ with the agent's observation of the environment $s_t$ to get the combined observation $\tilde{s}_t$.

\noindent\textbf{Min-Assumptions: Incorporating user control via raw action embedding.} In this setting, which we use in the majority of our experiments, we do not use any explicit goal inference. Instead, the policy directly takes in the user's actions $a^h_t$ and must learn to implicitly decode the user's intent and perform the task.\footnote{In principle, the user's past actions are also informative of intent, and a recurrent policy could effectively integrate these. In practice, we found a reactive policy to be more effective for our tasks.} To our agent, the user is part of the external environment, and the user's control is yet another source of observations, much like the output of any of the agent's other sensors. Because deep neural networks are end-to-end trainable, our agent can discover arbitrary relationships between user controls and observations of the physical environment, rather than explicitly assuming the existence of a goal. Our method jointly embeds the agent's observation of the environment $s_t$ with the user's control input $a^h_t$ by simply concatenating them, henceforth referred to as ``raw action embedding.'' In this setting, we set $u_t = a^h_t$.

\subsection{Q-Learning with User Control}

Model-free reinforcement learning with a human in the loop poses two challenges: (1) maintaining informative user input
and (2) minimizing the number of interactions with the environment. (1) If the user input is a suggested control, consistently ignoring the suggestion and taking a different action can degrade the quality of user input, since humans rely on feedback from their actions to perform real-time control tasks~\cite{laskey2017comparing}.
Additionally, some user policies may already be approximately optimal and only require fine-tuning. (2) Many model-free reinforcement learning algorithms require a large number of interactions with the environment, which may be impractical for human users.
To mitigate these two issues, we use deep Q-learning~\cite{watkins1992q} to learn an approximate state-action value function that can be used to select and evaluate actions. Specifically, we implement neural fitted Q-iteration (NFQI)~\cite{riedmiller2005neural} with experience replay~\cite{lin1993reinforcement}, a periodically updated target network~\cite{mnih2015human}, and double Q-learning~\cite{van2016deep}. This gets around a practical problem with using vanilla deep Q-networks (DQN)~\cite{mnih2015human} for human-in-the-loop learning: DQN performs a gradient update after each step, which can cause the task interface to lag and disrupts human control, whereas NFQI only performs gradient updates at the end of each episode. We chose Q-learning because (a) it is an off-policy algorithm, so we do not need to exactly follow the agent's policy and can explicitly trade off control between the user and agent, and (b) off-policy Q-learning tends to be more sample-efficient than policy gradient and Monte Carlo value-based methods~\cite{hessel2017rainbow}.

\subsection{Control Sharing}
Motivated by the discussion of (1) and (a)
in the previous section, we use the following behavior policy to select actions during and after Q-learning: select a feasible action closest to the user's suggestion, where an action is feasible if it isn't that much worse than the optimal action. Formally,
\begin{equation} \label{eq:beh-pol}
\pi_{\alpha}(a \mid \tilde{s}, a^h) = \delta\left(a = \argmax_{\{a : Q'(\tilde{s}, a) \geq (1 - \alpha) Q'(\tilde{s}, a^\ast)\}} f(a, a^h)\right),
\end{equation}
where $f$ is an action-similarity function and $Q'(\tilde{s}, a) = Q(\tilde{s}, a) - \min_{a' \in \mathcal{A}} Q(\tilde{s}, a')$ maintains a sane comparison for negative Q values: if $Q(\tilde{s}, a) < 0 ~ \forall a$ and $0 < \alpha < 1$, then the set of feasible actions would be empty if we didn't subtract a baseline from the Q values. The constant $\alpha \in [0, 1]$ is a hyperparameter that controls the tolerance of the system to suboptimal human suggestions, or equivalently, the amount of assistance. The functional form of the action feasibility condition is motivated by the fact that it is invariant to affine scaling of Q values.
The overall algorithm is summarized in Algorithm~\ref{alg:hitl-dql}.

\begin{algorithm}[t]
\small
\begin{algorithmic}
\State Initialize experience replay memory $\mathcal{D}$ to capacity $N$
\State Initialize $Q$-function with random or pretrained weights $\theta$
\State Initialize target action-value function $\hat{Q}$ with weights $\theta^{-} = \theta$
\For{episode $=1,M$}
\For {$t=1,T$}
	\State Sample action $a_t \sim \pi_{\alpha}(a_t \mid \tilde{s}_t, a^h_t)$ using equation~\ref{eq:beh-pol}
	\State Execute action $a_t$ and observe $(\tilde{s}_{t+1},a^h_{t+1},r_t)$
	\State Store transition $\left(\tilde{s}_t,a_t,r_t,\tilde{s}_{t+1}\right)$ in $\mathcal{D}$
	\If {$\tilde{s}_{t+1}$ \text{ is terminal}}
	\For {$k=1$ to $K$} \Comment{training loop}
	\State Sample minibatch $\left(\tilde{s}_j,a_j,r_j,\tilde{s}_{j+1}\right)$ from $\mathcal{D}$
	\State $y_j = r_j + \gamma \hat{Q}(\tilde{s}_{j+1}, \argmax_{a'} Q(\tilde{s}_{j+1}, a'; \theta); \theta^{-})$
	\State $\theta \leftarrow \theta - \eta\nabla_\theta \sum_{j} \left(y_j - Q(\tilde{s}_j, a_j; \theta) \right)^2$
	\EndFor
	\EndIf
	\State Every $C$ steps reset $\hat{Q} = Q$
\EndFor
\EndFor
\end{algorithmic}
\caption{Human-in-the-loop deep Q-learning}
\label{alg:hitl-dql}
\end{algorithm}

\section{Simulation Experiments}

\begin{figure*}[t]
    \centering
    \includegraphics[width=0.24\linewidth]{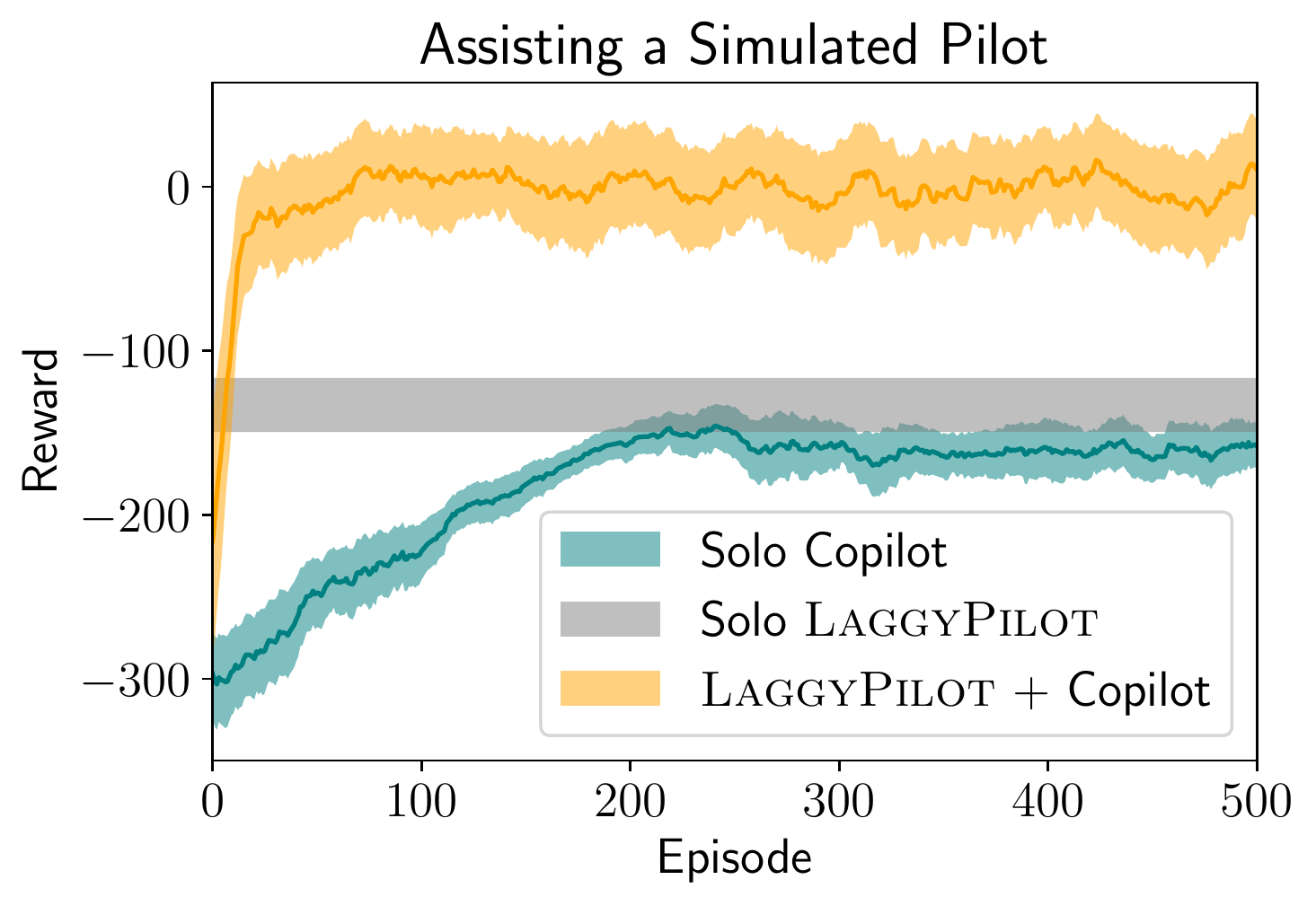}
    \includegraphics[width=0.24\linewidth]{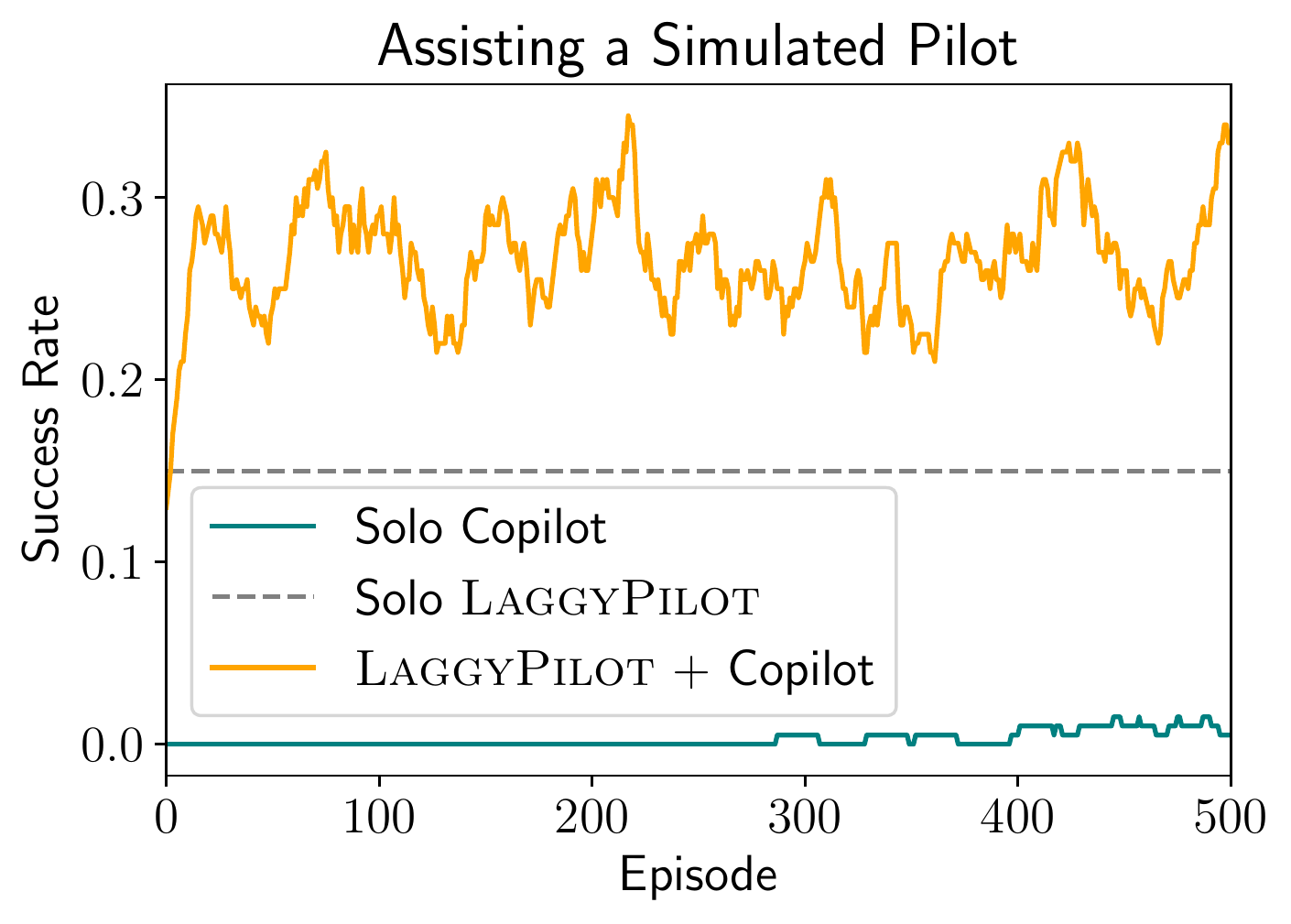}
    \includegraphics[width=0.24\linewidth]{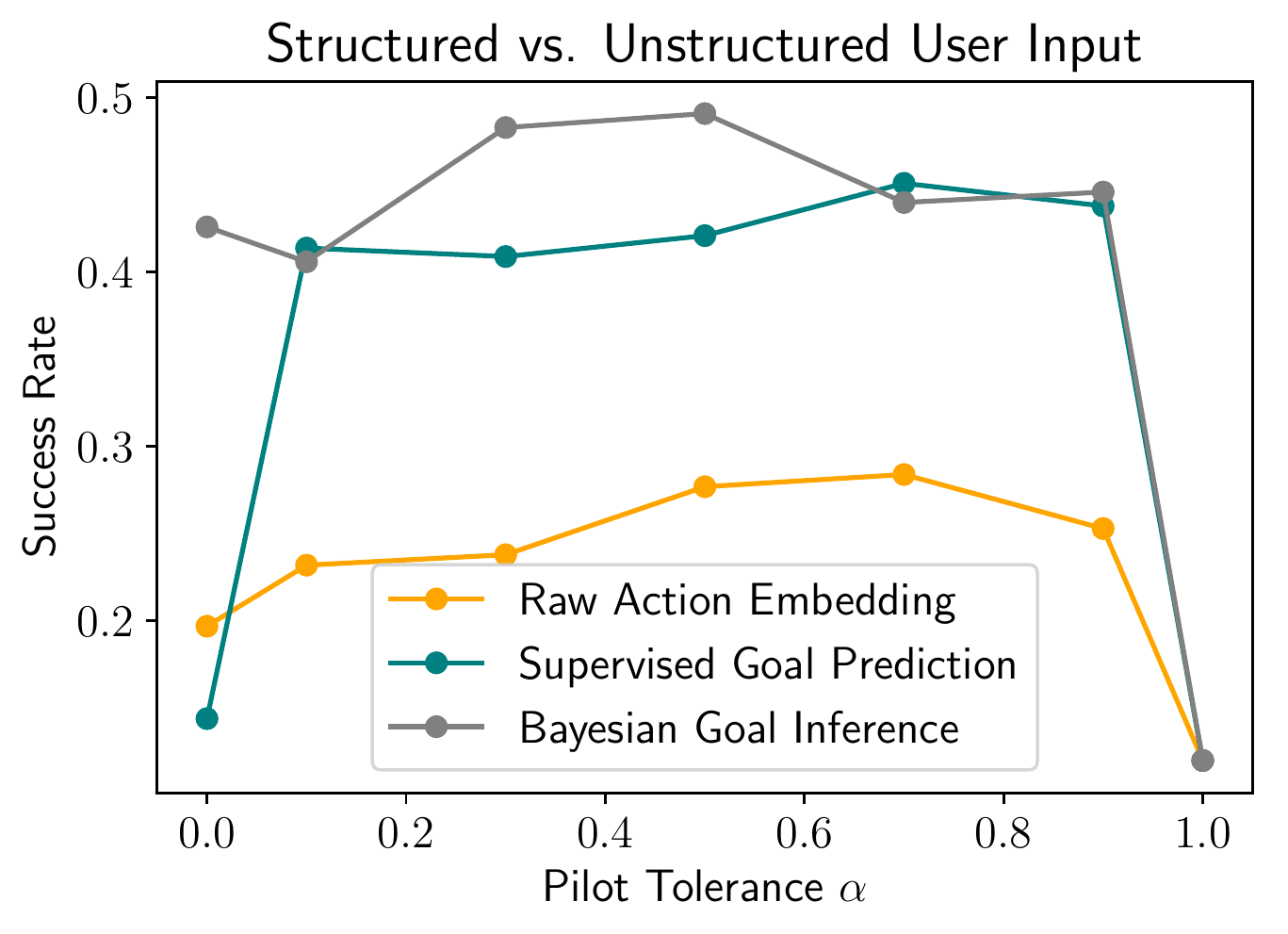}
    \includegraphics[width=0.24\linewidth]{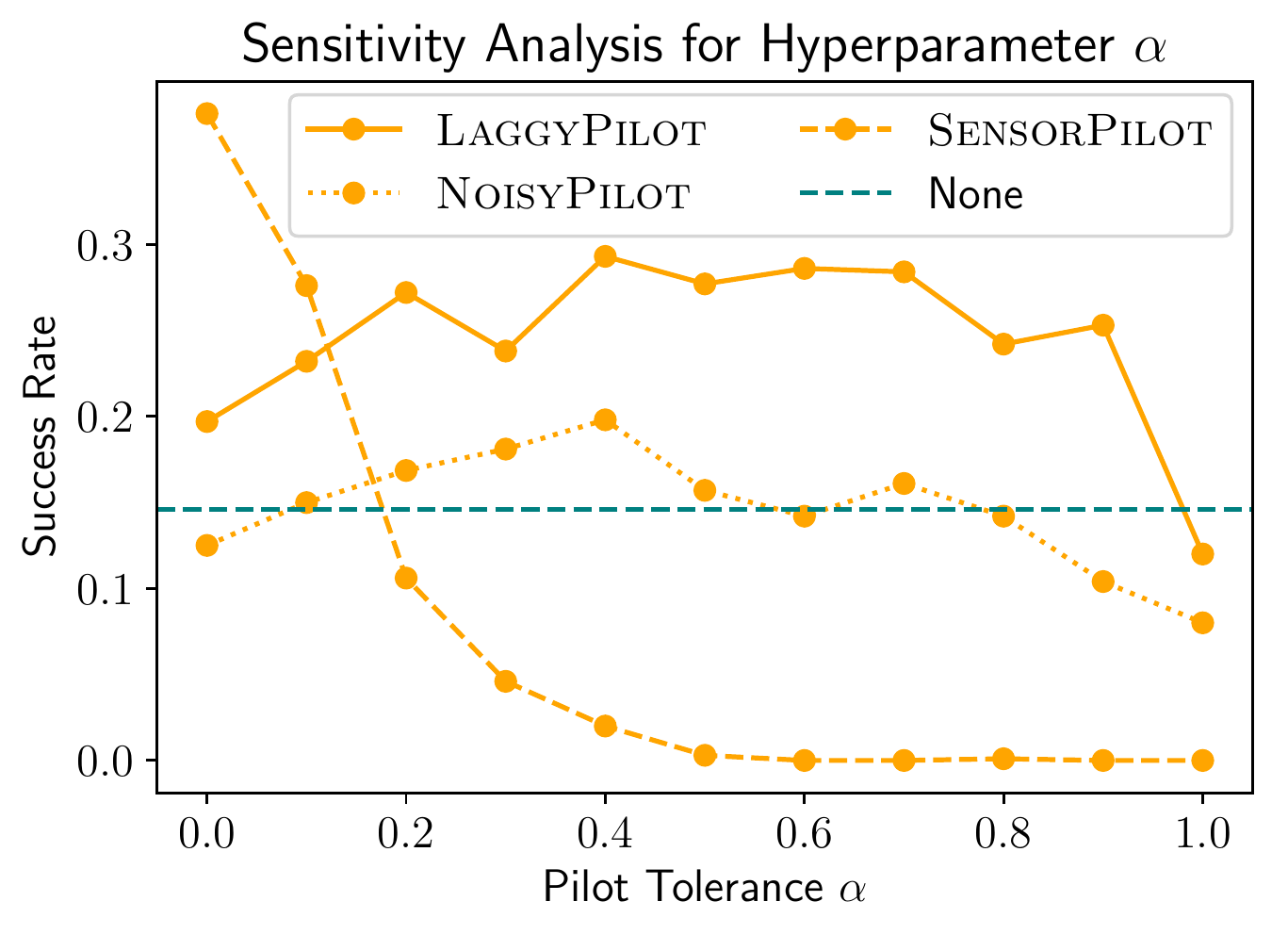}
    \caption{ (1,2) A copilot that leverages input from the synthetic \textsc{LaggyPilot} outperforms the solo \textsc{LaggyPilot} and solo copilot. The colored bands illustrate the standard error of rewards and success rates for ten different random seeds. Rewards and success rates are smoothed using a moving average with a window size of 20 episodes. (3) The benefit of using Bayesian goal inference or supervised goal prediction depends on $\alpha$. Each success rate is averaged over ten different random seeds and the last 100 episodes of training. (4) The effect of varying $\alpha$ depends on the user model. Each success rate is averaged over ten different random seeds and the last 100 episodes of training.}
    \label{fig:laggy-pi-results}
\end{figure*}

\begin{table*}[t]
  \caption{Evaluation of simulated pilot-copilot teams on Lunar Lander. Rewards are shown with their standard error on ten different random seeds and the last 100 episodes of copilot training for teams with a copilot; on 100 episodes for teams without a copilot.}
  \label{all-results-no-gds}
  \centering
  \begin{tabular}{lllllllll}
    \toprule
    & \multicolumn{3}{c}{Without Copilot} &  \multicolumn{3}{c}{With Copilot}                   \\
    \cmidrule(lr){2-4} \cmidrule(lr){5-9}
    Pilot & Reward & Success Rate & Crash Rate & Reward & Success Rate & Crash Rate & Training Episodes & $\alpha$ \\
    \midrule
    None & - & - & - & $-151 \pm 17$ & 0.026 & 0.156 & 742 & 0.0 \\
    Sensor & $-479 \pm 14$  &  0.000  & 1.000  & $-82 \pm 0$ & 0.060 & 0.650 & 800 & 0.2 \\
    Laggy & $-133 \pm 16$  &  0.150  & 0.750  & $8 \pm 30$ & 0.287 & 0.186 & 236 & 0.8 \\
    Noisy & $-72 \pm 8$  &  0.150  & 0.700  & $-28 \pm 0$ & 0.240 & 0.160 & 604 & 0.5 \\
    Optimal & $75 \pm 6$  &  0.720  & 0.030  & - & - & - & - & - \\
    \bottomrule
  \end{tabular}
\end{table*}

We begin our experiments with an analysis of our method under different simulated users. To simplify terminology, we henceforth refer to the user as the \emph{pilot} and the semi-autonomous agent as the \emph{copilot}. Our central hypothesis is that our method can improve a pilot's performance despite not knowing the world's dynamics and the pilot's policy, or assuming a particular set of goals. Simulating pilots enables us to take a deeper dive into different aspects of our method (like the effects of the tolerance parameter $\alpha$, and of training and testing on different types of input) before testing on real users -- after all, simulated pilots do not run out of patience.

\noindent\textbf{The Lunar Lander System.} \label{lander-spec}
We use the Lunar Lander game from OpenAI Gym~\cite{1606.01540} (see the bottom-left panel of Figure \ref{fig:front-fig}) as our test platform for this part of our experiments.
The objective of the game is to pilot the lunar lander vehicle to a specified landing site on the ground without crashing using two lateral thrusters and a main engine. Each episode lasts at most 1000 steps, and runs at 50 frames per second. An episode ends when the lander crashes, flies out of bounds, remains stationary on the ground, or time runs out. The action space $\mathcal{A}$ consists of six discrete actions that correspond to the \{left, right, off\} steering commands and \{on, off\} main engine settings. The state $s \in \mathbb{R}^8$ is an eight-dimensional vector that encodes the lander's position, velocity, angle, angular velocity, and indicators for contact between the legs of the vehicle and the ground. The x-coordinate of the landing site is selected uniformly at random at the beginning of each episode, and is not directly accessible to the agent through the state $s$. A human playing the game can see two flags demarcating the landing site, and can supply a suggested control $a^h \in \mathcal{A}$ -- depending on the user policy, $a^h$ could be an approximately-optimal action, a signal that encodes the relative direction of the landing site, etc. Thus, in order to perform the task, the agent needs to leverage $a^h$ to maneuver toward the landing site.

The agent uses a multi-layer perceptron with two hidden layers of 64 units each to approximate the Q function $\hat{Q} : \mathcal{S} \times \mathcal{A}^2 \to \mathbb{R}$. The action-similarity function $f(a, a^h)$ in the agent's behavior policy counts the number of dimensions in which actions $a$ and $a^h$ agree (e.g., $f((\text{left}, \text{on}), (\text{left}, \text{off})) = 1$).
As discussed earlier in Section \ref{mfsa}, the agent's reward function is composed of a hard-coded function $R_{\text{general}}$ and a user-generated signal $R_{\text{feedback}}$. $R_{\text{general}}$ penalizes speed and tilt, since moving fast and tipping over are generally dangerous for any pilot regardless of their intent. $R_{\text{feedback}}$ emits a large positive reward at the end of the episode if the vehicle successfully lands at the intended site, or a large negative reward if it crashes or goes out of bounds.

\subsection{Testing Unstructured Copilot Performance} \label{incomp-task-spec}
We now test the central hypothesis that our method, model-free shared autonomy, improves a pilot's performance. We do this first in the Min-Assumptions setting, where the dynamics, user policy, and goal space are all unknown. We then test our ability to leverage this information when it exists in the next section.

\noindent\textbf{Manipulated variables.}
We manipulate (1) the operator team composition: a solo pilot, a solo copilot, or our method -- a pilot assisted by a copilot; and (2) the policy followed by the simulated pilot -- a categorical variable that can take on four values: None (always executes a noop), \textsc{LaggyPilot}, \textsc{NoisyPilot}, and \textsc{SensorPilot}.

\textsc{LaggyPilot} is an optimal pilot except that it can't change actions quickly, which for a real human might be the result of poor reaction time.
The \textsc{LaggyPilot} policy is trained as follows: augment the state vector with the landing site coordinates, train a reinforcement learning agent using vanilla DQN, and corrupt the trained policy by forcing it to repeat the previously executed action with fixed probability $p = 0.85$. This causes each action to repeat for a number of steps that follows a geometric distribution.

\textsc{NoisyPilot} is an optimal pilot except that it occasionally takes the wrong action, which for a real human might be the result of mistakenly pressing the wrong key. It uses the same training procedure as \textsc{LaggyPilot} but follows an $\epsilon$-greedy behavior policy at test time ($\epsilon = 0.3$).

\textsc{SensorPilot} tries to move toward the landing site by firing the appropriate lateral thruster, but is oblivious to gravity and doesn't use the main engine; these actions provide enough signal for an assistive copilot to deduce the location of the landing site, which may be all the human is willing to do.

\noindent\textbf{Dependent measures.} We measure reward, success rate, and crash rate.

\noindent\textbf{Hypothesis.} We hypothesize that a pilot-copilot team with a simulated pilot will perform better on the Lunar Lander game than a solo pilot or solo copilot.

\noindent\textbf{Analysis.}
The results in Figure \ref{fig:laggy-pi-results} (first two plots) show that a copilot which leverages input from \textsc{LaggyPilot} outperforms the solo \textsc{LaggyPilot} and solo copilot: the combined pilot-copilot team crashes and goes out of bounds less often, uses less fuel, follows stabler trajectories, and finds the landing site more often than the other two solo teams. The solo copilot and combined pilot-copilot teams learn from experience, whereas the solo \textsc{LaggyPilot} is pretrained and frozen; hence the stationarity of the gray curve. Table \ref{all-results-no-gds} shows that \textsc{NoisyPilot} and \textsc{SensorPilot} also benefit from assistance, although \textsc{SensorPilot}'s success rate does not substantially increase.

To measure the sensitivity of the copilot's performance to the pilot tolerance hyperparameter $\alpha$ (recall Equation \ref{eq:beh-pol}), we sweep different values of $\alpha$ while shaping the reward $R_{\text{feedback}}$ to improve the performance of \textsc{SensorPilot}. The results in Figure \ref{fig:laggy-pi-results} (bottom right) show the effects of varying $\alpha$ for different simulated pilot models: $\alpha = 0$ is optimal for \textsc{SensorPilot}, and $\alpha \approx 0.5$ is optimal for \textsc{LaggyPilot} and \textsc{NoisyPilot}.

\subsection{The Benefit of Structure when Structure Exists}

In some tasks, the user's private information will indeed be a goal, and we will indeed know the set of candidate goals and the policy that user follows given a goal. In this section, we show the adaptions of our method for Known-Goal-Space and Known-User-Policy from Section \ref{mfsa} can effectively leverage this information when it exists.

\noindent\textbf{Manipulated variables.}
We manipulate (1) the input decoding mechanism -- a categorical variable that can take on three values: Bayesian goal inference, supervised goal prediction, and raw action embedding; and (2) the pilot tolerance $\alpha \in [0, 1]$ -- a continuous variable sampled uniformly across the unit interval.

\noindent\textbf{Hypothesis.} We hypothesize that a copilot that uses Bayesian goal inference or supervised goal prediction to interpret user control inputs from \textsc{LaggyPilot} will outperform a copilot that uses raw action embedding.

\noindent\textbf{Analysis.}
The results in Figure \ref{fig:laggy-pi-results} (third plot) show that when the goal space and user model are known, Bayesian goal inference and supervised goal prediction outperform raw action embedding. Bayesian goal inference enables much better assistance when the user model is approximately correct: \textsc{LaggyPilot} behaves similarly enough to an optimal pilot that maximum entropy inverse reinforcement learning generates high-accuracy estimates of the landing site.
As a result, Bayesian goal inference performs better than supervised goal prediction and raw action embedding on \textsc{LaggyPilot}.
We conclude that when an approximately correct user model is available, one should take advantage of it by using Bayesian goal inference instead of supervised goal prediction or raw action embedding. When the user model is unknown ex-ante, then one should use supervised goal prediction instead of raw action embedding.

\subsection{Adapting to Diverse Users}

Next, we investigate to what extent the copilot's learned policy is adapted to the pilot it assists at training time.
User-specific adaptation is important because it would enable our method to generalize to tasks in which users display a range of behavior policies with distinct types of errors that cannot simultaneously be corrected by a general assistance feature.

\noindent\textbf{Manipulated variables.}
We manipulate (1) the policy followed by the simulated pilot used to train the copilot and (2) the policy followed by the simulated pilot used to evaluate the copilot -- both categorical variables that can each take on four values: None (always executes a noop), \textsc{SensorPilot}, \textsc{LaggyPilot}, and \textsc{NoisyPilot}.

\noindent\textbf{Hypothesis.}
We hypothesize that the copilot learns an assistive policy that is personalized to the individual user, and that a copilot trained with one type of simulated pilot will perform better if evaluated with the same type of pilot than with a different pilot.

\begin{table}[t]
  \caption{Training and testing with different pilots on Lunar Lander. Success rates shown for 100 episodes.}
  \label{cross-eval}
  \centering

  \begin{tabular}{lllll}
    \toprule
     & \multicolumn{4}{c}{Evaluation Pilot} \\
    \cmidrule(lr){2-5}
    Training Pilot & None & Sensor & Laggy & Noisy \\
    \midrule
None & 0.02 & 0.02 & 0.29 & 0.04 \\
Sensor & 0.18 & 0.46 & 0.31 & 0.23 \\
Laggy & 0.00 & 0.00 & 0.31 & 0.23 \\
Noisy & 0.10 & 0.10 & 0.38 & 0.21 \\
    \bottomrule
  \end{tabular}
\end{table}

\noindent\textbf{Analysis.}
The results in Table \ref{cross-eval} hint that the copilot trained to assist \textsc{SensorPilot} acquires a relatively unique assistive policy, and that assisting \textsc{SensorPilot} requires something qualitatively different than assisting other pilots. A copilot trained with \textsc{SensorPilot} does not help other simulated pilots as well as it helps \textsc{SensorPilot}. Copilots trained with non-\textsc{SensorPilot} pilots do not assist \textsc{SensorPilot} as well as a copilot trained with \textsc{SensorPilot}. In contrast, a copilot evaluated with \textsc{LaggyPilot} or \textsc{NoisyPilot} performs equally well when trained with either of those two pilots. These results may be explained by the fact that \textsc{SensorPilot} implements a goal-signaling policy that is qualitatively distinct from \textsc{LaggyPilot} and \textsc{NoisyPilot}, which both implement policies based on perturbations of an optimal pilot.

Another takeaway from Table \ref{cross-eval} is that a copilot trained without a pilot learns an assistive policy that is just as effective at helping \textsc{LaggyPilot} as a copilot policy trained with \textsc{LaggyPilot} in the loop. This suggests that the copilot can still learn useful assistive behaviors even when there is no pilot in the loop during training. Furthermore, it may enable us to save human pilots time by pretraining the copilot without the human pilot, then fine-tuning the pretrained copilot with the human pilot.

\section{User Study with a Game Agent}

\begin{figure*}[t]
    \centering
    \includegraphics[width=0.24\linewidth]{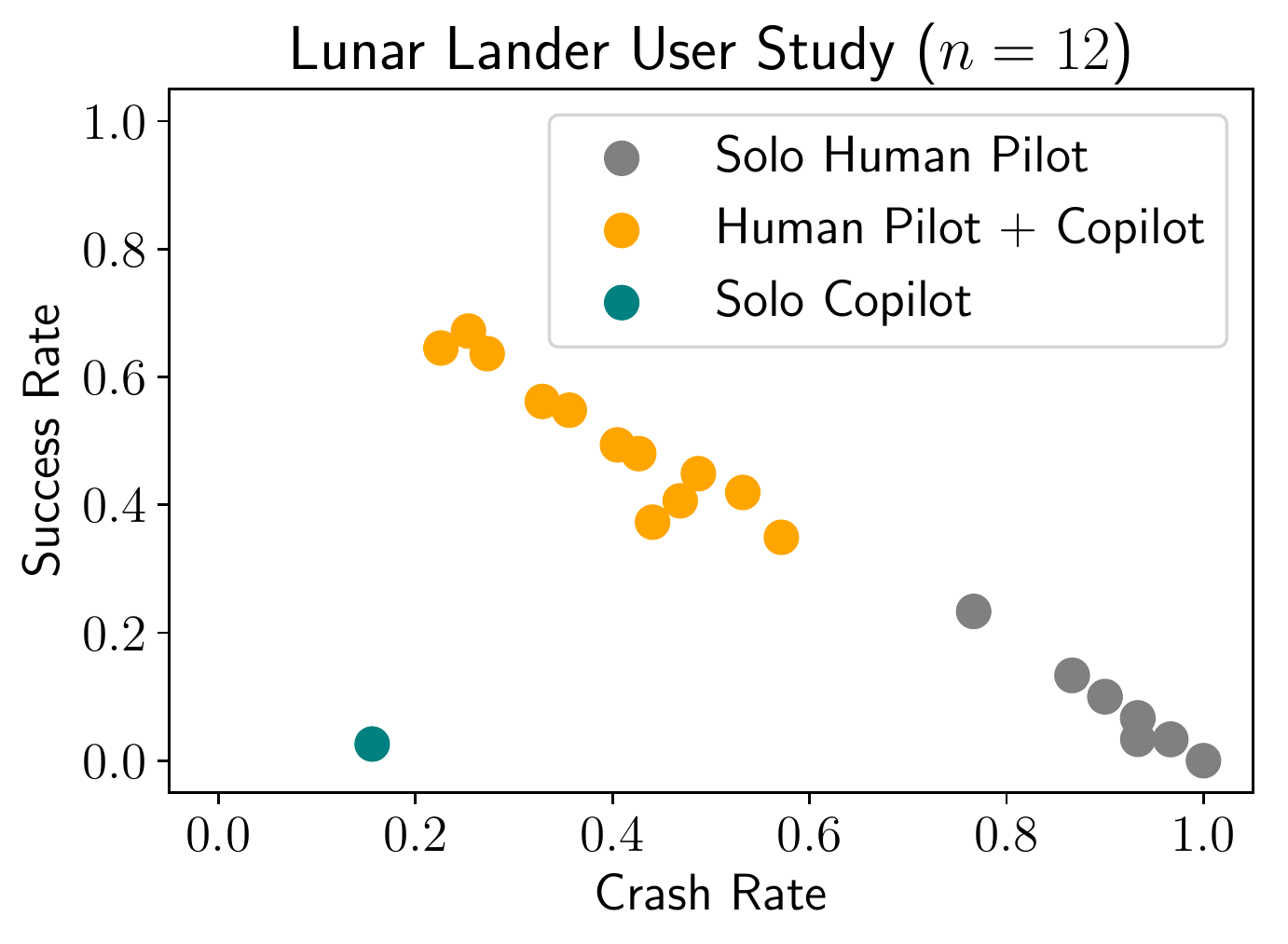}
    \includegraphics[width=0.24\linewidth]{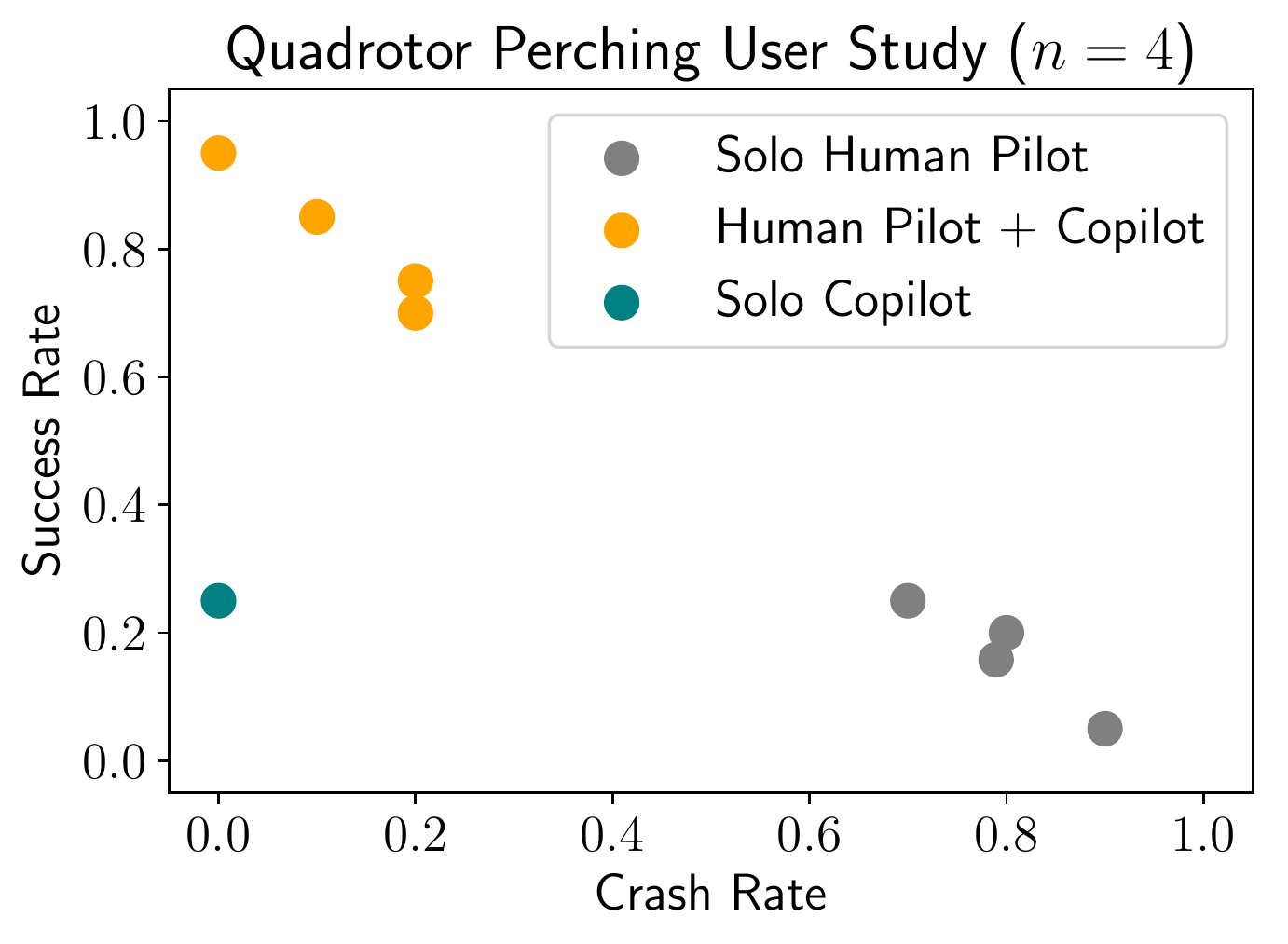}
    \includegraphics[width=0.24\linewidth]{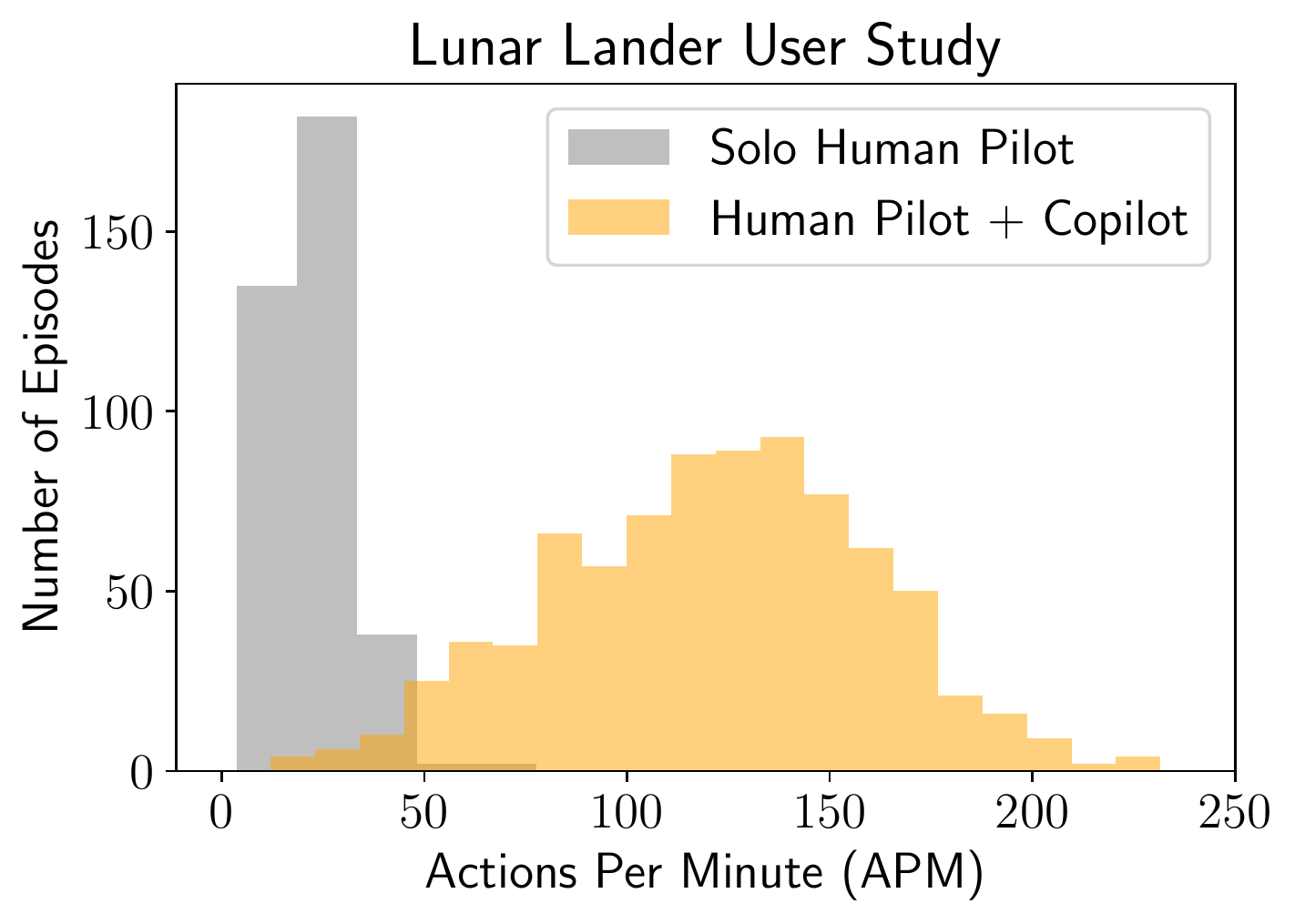}
    \includegraphics[width=0.24\linewidth]{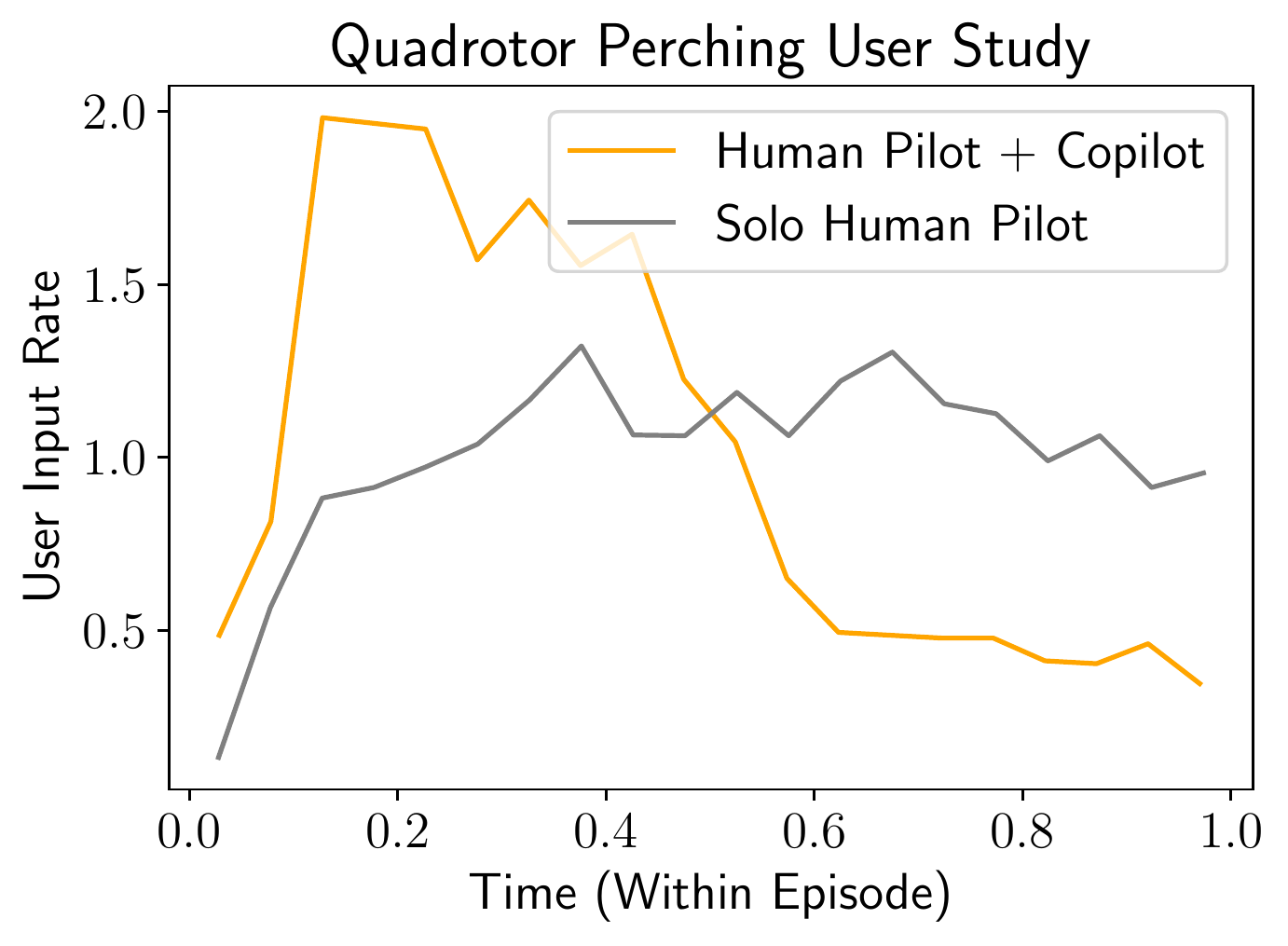}
    \caption{(1) Evaluation of real humans on Lunar Lander. Success and crash rates averaged over 30 episodes for teams with human pilots. (2) Evaluation of real humans on the quadrotor perching task. Success and crash rates averaged over 20 episodes. (3) Pilot-copilot teams in the Lunar Lander game are able to switch between actions more quickly than solo human pilots, which enables them to better stabilize flight. (4) On their own, users tend to provide input at a constant rate throughout an episode. When assisted by a copilot, users initially rotate the drone to orient the camera at the target object, then defer to the copilot to fly to the landing pad.}
    \label{fig:user-exp}
\end{figure*}

We saw in simulation that our method can improve the performance of different kinds of pilots. Next, we test whether it can actually help real people in a teleoperation task.

\noindent\textbf{Manipulated variables.}
We manipulated the team structure: solo copilot, solo human pilot, and our method -- human pilot with a copilot. We use the same Lunar Lander environment for this part of the experiment. $R_{\text{feedback}}$ is a terminal reward as before.

\noindent\textbf{Dependent measures.}
Our objective measures are success rate and crash rate. We additionally introduce some informal subjective measures, where in each condition we ask participants about their experience, to help us understand their perception of the copilot.

\noindent\textbf{Hypothesis.} We hypothesize that a pilot-copilot team with a real human pilot will perform better than a solo human pilot or solo copilot.

\noindent\textbf{Subject allocation.}
We recruited 11 male and 1 female participants, with an average age of 24. Each participant was provided with the rules of the game and a short practice period of 20 episodes to familiarize themselves with the controls and dynamics.
To avoid the confounding effect of humans learning to play the game better over time, we counterbalanced the order of the two conditions that required human play (solo pilot, and assisted pilot). Each condition lasted 30 episodes.

To speed up learning, the copilot was pretrained without a pilot in the loop then fine-tuned on data collected from the human pilot. Pilot tolerance $\alpha = 0.6$ was chosen heuristically to match the difficulty of the game and the average human user's skill level. The default game environment is too challenging for human pilots, so it was modified to make the vehicle's legs more resistant to crashing on impact with the ground. Additionally, pilot tolerance was set to $\alpha = 0$ when the human was not pressing any keys, and an additional key was introduced to enable the user to explicitly enter a noop input with $\alpha = 0.6$.

\noindent\textbf{Analysis.}
Figure \ref{fig:user-exp} (first plot) shows a clear quantitative and qualitative benefit to combining a real human pilot with a copilot. Humans follow a tortuous path with sudden drops and difficult course corrections, leading to fewer successes and more crashes. With a copilot, the human follows a smooth, gradual descent to the landing site, leading to significantly more successes and significantly fewer crashes than without a copilot for each of the participants. We ran a repeated measures ANOVA with the presence of the copilot as a factor influencing success and crash rates, and found that $f(1,11)=165.0001, p<0.0001$ for the success rate and $f(1,11)=259.9992, p<0.0001$ for the crash rate. The combined human pilot-copilot team succeeds significantly more often than the solo copilot, at the expense of crashing significantly more often. For each of the participants, we ran a binomial test comparing their success rate and crash rate in the combined pilot-copilot team to those of the solo copilot and found that $p < 0.01$ for all comparisons.

The subjective evaluations generally suggest that users benefited from the copilot. The assistive system was particularly helpful in avoiding crashing, but perceived to be somewhat inconsistent in its behavior and too aggressive in stabilizing flight at the expense of slowing down the lander's descent.

\section{User Study with a Physical Robot: Quadrotor Perching}

One of the drawbacks of analyzing Lunar Lander is that the game interface
and physics do not reflect the complexity and unpredictability of a real-world robotic shared autonomy task.
To evaluate our method in a more realistic environment, we formulate a ``perching'' task for a real human flying a real quadrotor: land the vehicle on a level, square landing pad at some distance from the initial take-off position, such that the drone's first-person camera is pointed at a specific object in the drone's surroundings, without flying out of bounds or running out of time. Perching a drone at an arbitrary vantage point enables it to be used as a mobile security camera for surveillance applications. Humans find it challenging to simultaneously point the camera at the desired scene and navigate to the precise location of a feasible landing pad under time constraints.
An assistive copilot has little trouble navigating to and landing on the landing pad, but does not know where to point the camera because it does not know what the human wants to observe after landing. Together, the human can focus on pointing the camera and the copilot can focus on landing precisely on the landing pad.

\noindent\textbf{Robot task.}
Figure \ref{fig:front-fig} (b, c) illustrates the experimental setup. We fly the Parrot AR-Drone 2 in an indoor flight room equipped with a Vicon motion capture system to measure the position and orientation of the drone as well as the position of the landing pad. Users are only allowed to look through the drone's first-person camera to navigate, and are blocked from getting a third-person view of the drone. Each episode lasts at most 30 seconds. An episode begins when the drone finishes taking off. An episode ends when the drone lands, flies out of bounds, or time runs out. The action space $\mathcal{A}$ consists of 18 discrete actions that correspond to moving left, right, forward, back, descending, or hovering in place and simultaneously rotating (yawing) clockwise, counter-clockwise, or not rotating. The state $s \in \mathbb{R}^{10}$ is a ten-dimensional vector that encodes the vehicle's position, velocity, angle, angular velocity, and the horizontal components of the difference between the landing pad position and the vehicle's position.
At the beginning of each episode, the starting position and orientation of the drone are randomized and the user is told that their goal is to point the camera at an object selected randomly from a set of four in the vicinity: a red chair, a gray chair, white styrofoam boards, or a door. The agent's state does not include this target orientation, which is necessary for success. Success is defined as landing on the pad (evaluated automatically using motion tracking) while orienting the camera at the correct object, which is evaluated by the human experimenter with a button press at the end of the episode.
Crashing is defined as landing outside the landing pad or going out of bounds.

As before, the agent uses a multi-layer perceptron with two hidden layers of 64 units each to approximate the Q function $\hat{Q} : \mathcal{S} \times \mathcal{A}^2 \to \mathbb{R}$. The action-similarity function $f(a, a^h)$ in the agent's behavior policy counts the number of dimensions in which actions $a$ and $a^h$ agree (e.g., $f((\text{left}, \text{rotate clockwise}), (\text{left}, \text{rotate counter-clockwise})) = 1$).
As discussed earlier in Section \ref{mfsa}, the agent's reward function is composed of a hard-coded function $R_{\text{general}}$ and a user-generated signal $R_{\text{feedback}}$. $R_{\text{general}}$ penalizes distance from the landing pad, since moving toward the pad is generally useful to all pilots regardless of their desired camera orientation. $R_{\text{feedback}}$ emits a large positive reward at the end of the episode if the task was completed successfully, or a large negative reward in the event of a crash.

\noindent\textbf{Manipulated variables.}
We manipulate the pilot-copilot team membership as before.

\noindent\textbf{Dependent measures.}
Performance is measured using the dependent factors of success rate and crash rate. As before, we ask participants about their experience (see Table 2 in the supplementary material) to help us understand their perception of the copilot.

\noindent\textbf{Hypothesis.} We hypothesize that a pilot-copilot team with a real human pilot will perform better on the quadrotor perching task than a solo human pilot or a solo copilot.

\noindent\textbf{Subject allocation.}
We recruited 3 male and 1 female participants, with an average age of 23. Each participant was provided with the rules of the game and a short practice period of 2 episodes to familiarize themselves with the controls and dynamics.
To avoid the confounding effect of humans learning to play the game better over time, we counterbalanced the order of the two conditions that required human play (solo pilot, and assisted pilot). Each condition lasted 20 episodes.

To speed up learning, the copilot was pretrained
in simulation without a pilot in the loop then fine-tuned on data collected from the human pilot. The pretraining simulation assumed an idealized physics model in which the drone is a point mass, there are no external forces, linear velocity commands are executed without any noise, and sensors have zero measurement error. In the pretraining simulation, a target angle (yaw) is randomly sampled for each episode to simulate the random choice of a target object for the camera in the real world. As before, the agent cannot directly access this target angle through its state.

With a real human pilot in the real world, pilot tolerance was set to $\alpha = 0$ when the human was not pressing any keys, and otherwise set to $\alpha = 1$.

\noindent\textbf{Analysis.}
Figure \ref{fig:user-exp} (second plot) shows a clear quantitative and qualitative benefit to combining a real human pilot with a copilot. Humans are rarely able to arrive at the landing pad, leading to fewer successes and more crashes. With a copilot, the human consistently gets to the landing pad, leading to significantly more successes and significantly fewer crashes than without a copilot. The sample size of $n = 4$ participants is relatively small, so the evidence is mainly anecdotal and should be interpreted in the context of the larger simulation experiments and user study on the Lunar Lander game. With that in mind, we ran a repeated measures ANOVA with the presence of the copilot as a factor influencing success and crash rates, and found that $f(1,3)=44.1045, p<0.01$ for the success rate and $f(1,3)=62.3151, p<0.01$ for the crash rate. The combined human pilot-copilot team succeeds significantly more often than the solo copilot, at the expense of crashing significantly more often. For each of the participants, we ran a binomial test comparing their success rate and crash rate in the combined pilot-copilot team to those of the solo copilot and found that $p < 0.01$ for all comparisons.
The subjective evaluations in Table 2 of the supplementary material generally suggest that users benefited from the copilot.

\section{Discussion}

In this paper, we contribute an algorithm for shared autonomy that uses model-free reinforcement learning to help human users with tasks with unknown dynamics, user policies, and goal representations. We introduce a behavioral policy for deep Q-learning that enables users to directly control the level of assistance, as well as a decomposition of the reward function that enables the system to quickly learn generally useful behaviors and also adapt to individual users. Our user studies with a virtual agent and a real robot suggest that this method can indeed be effective at improving user performance.

Several weaknesses and open questions remain to be addressed.
Inferring user intent in general will require memory. Several existing techniques may accomplish this, including concatenating $m$ previous frames with the current observation, or adding recurrent connections to the copilot policy architecture as in~\cite{hausknecht2015deep}.
Finally, users will adapt to the robot's interface, and explicitly capturing this may improve copilot training and inform theoretical guarantees on convergence~\cite{nikolaidis2017human}.

\section{Acknowledgements}

We would like to thank Oleg Klimov for open-sourcing his implementation of the Lunar Lander game, which was originally developed by Atari in 1979. This work was supported in part by a Berkeley EECS Department Fellowship for first-year Ph.D. students, Berkeley DeepDrive, computational resource donations from Amazon, NSF IIS-1700696, and AFOSR FA9550-17-1-0308.

\bibliographystyle{plainnat}
\bibliography{master}

\end{document}